# Machine Translation Systems in India

**Sugata Sanyal**
Corporate Technology Office,
Tata Consultancy Services,
Mumbai, India
sugata.sanyal@tcs.com

**Rajdeep Borgohain (*corresponding author*)**
Department of Computer Science and Engineering,
Dibrugarh University Institute of Engineering and Technology, Dibrugarh, India
rajdeepgohain@gmail.com

**Abstract —** *Machine Translation is the translation of one natural language into another using automated and computerized means. For a multilingual country like India, with the huge amount of information exchanged between various regions and in different languages in digitized format, it has become necessary to find an automated process from one language to another. In this paper, we take a look at the various Machine Translation System in India which is specifically built for the purpose of translation between the Indian languages. We discuss the various approaches taken for building the machine translation system and then discuss some of the Machine Translation Systems in India along with their features*

## I. INTRODUCTION

Machine Translation is the automated process of translating one natural language to another. With the perspective of Indian languages, Machine Translation holds tremendous potential in various domains like health, education, information technology, business and various governmental agencies. India being a multi lingual country, languages vary from region to region. At present there are a total of 22 official languages in India. With the advent of Information Technology, many digitized documents, web pages are coming up in local languages, and it has become indispensable to build systems that would serve the purpose of translating within the Indian languages. Manual translation of these documents is not only time consuming but also has a tremendous cost factor involved. As such coming up with machine translation systems that translate between the Indian languages has become very important. Moreover, translating these regional languages to English languages is also necessary. Many researchers have started working on Machine Translation systems specifically catered for the Indian languages and have gained very satisfactory results.

The research scenario in India is relatively young and machine translation gained momentum in India only from 1980 onwards with institutions like IIT Kanpur, IIT Bombay IIIT Hyderabad, University of Hyderabad, NCST Mumbai, The Technology Development in Indian Languages (TDIL), and CDAC Pune playing a major role in developing the systems [17, 18]. Since then, many Machine Translation systems have been developed in India which has used different approaches for translating between the languages. Here, we take a look at the major Machine Translation systems which have been used for translation between the Indian languages.

The rest of the paper is organized in the following way: Section 2 gives an idea of the different approaches to build a machine translation system. Section 3 discusses the direct approach for building the machine translation system. In section 4 we discuss the Rule Based system. Section 5 gives an idea of the corpus based approaches. In section 6 we take a look at various Machine Translation Systems in India along with their features and we finally conclude the paper in the next section.

## II. APPROACHES IN MACHINE TRANSLATION

The process of Machine Translation can be broadly classified into the following approaches Direct Machine Translation, Rule Based Machine Translation, Corpus Based Machine Translation.

The Rule Based Approach can be further subdivided into Transfer Based approach and Interlingua approach while the Corpus based approach can be categorized into Statistical Machine Translation and Example based approach.

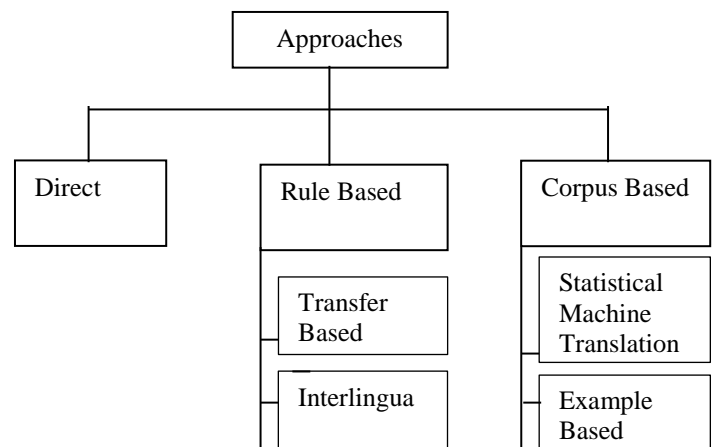

Figure 1 Machine Translation Approaches

## III. DIRECT MACHINE TRANSLATION

Direct Machine Translation is the one of the simplest machine translation approach. In Direct Machine Translation, a direct word by word translation of the input source is carried out with the help of a bilingual dictionary and after which some syntactical rearrangement are made.

In Direct Machine Translation a language called the source language is given as input and the output is known as the target language. Typically, the approach is unidirectional and only takes one language pair into consideration at a time.

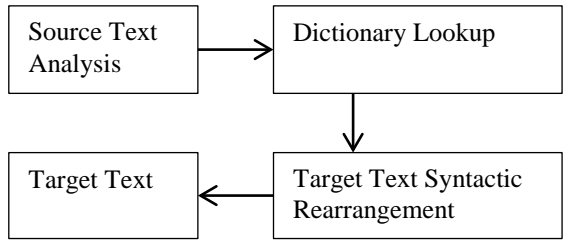

Figure 2 Direct Machine Translation Systems

The most well-known Direct Machine Translation system for Indian languages are Anusaaraka [10], developed by IIT Kanpur which is now being continued at IIT Hyderabad, Hindi to Punjabi Machine Translation System [5] developed at Punjab University which achieved an accuracy of 95 %.

## IV. RULE BASED MACHINE TRANSLATION

The Rule Based Machine Translation System takes into account semantic, morphological and syntactic information from a bilingual dictionary and grammar and based on these rules generate the output target language from the input source language by producing an intermediate representation. Rule based system is further classified as Interlingua based Machine Translation and Transfer based Machine Translation depending on the intermediate representation.

### A. Interlingua Approach

The Interlingua Approach converts words into an intermediate language IL, which is typically a universal language created for the system to use it as an intermediate for translation into more than one target language. The Interlingua approach has an analyzer which produces the intermediate representation of the source language. The synthesizer takes over from the analyzer and produces target sentences given by the analyzer

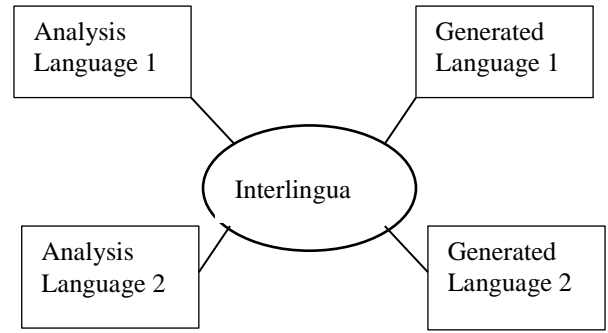

Figure 3 Interlingual Machine Translation

The major Machine Translation system in India based on Interlingua approach are Anglabharti [3], developed at IIT Kanpur which uses an intermediate structure Pseudo Lingua for Indian Languages. AnglaHindi [6] is an extension of Anglabharti which uses rule-bases, example-base and statistics to obtain translation for frequently encountered noun and verb phrasal. Another notable system is the Universal Natural Language English to Hindi Machine Translation system [7] developed at IIT Bombay which uses a universal natural language as propose by United Nations University. Currently work is in progress for English to Marathi and English to Bengali UNL system.

### A. Transfer based approach

The Transfer based approach uses translation rules to translate the input language to the output language which is done in three phases. The approach uses a dictionary to directly convert source into target whenever a sentence matches one of the transfer rules. Source language dictionary, Target language dictionary and a bilingual dictionary is used for this purpose. The translation mechanism is carried out in three phases. Firstly, the source language is converted into intermediate representation which is subsequently converted into target language representation in the second phase. The third phase involves generation of the final target language.

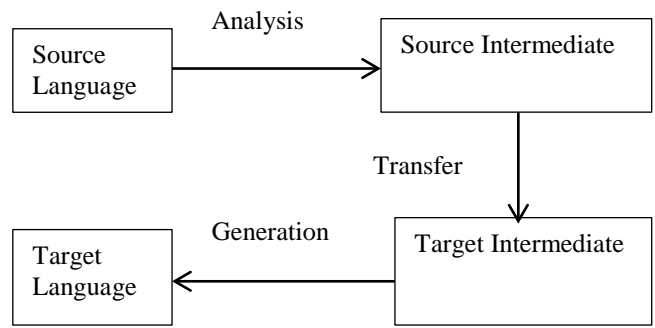

Figure 4 Transfer Based Machine Translation

Indian Machine Translation system based on Transfer based

approach includes Mantra, developed by Applied Artificial Intelligence group of CDAC, Pune. Another translation system is the Matra system also developed by CDAC, Pune [11]. Other systems include Shakti Machine Translation system [13] developed by IIIT Hyderabad and IISC Bangalore which works by combining rule based and statistical approach. Anubaad [12] is another translation system which is a hybrid system which uses n-gram approach for POS tagging and works at sentence level developed at CDAC, Kolkata. Another notable system is the English to Kannada Machine Translation system [11], developed at University of Hyderabad which uses Universal Clause Structure Grammar Formalism. Sampark [11] is another translation system which translates Indian Languages to Indian Languages Machine developed at IIIT Hyderabad which was based on Paninian framework. Recently, researchers have introduced distributed computing [2] and cloud computing [15, 16] for Indian machine translation systems. The Sampark system is deployed in cloud environment which reduces the deployment time [19] and distributed system environment which increases the throughput [22].

## V. CORPUS BASED MACHINE TRANSLATION

Corpus based machine translation have evolved as the preferred approach for machine translation. In this approach, a bilingual text corpus is taken and trained to get the desired output. The corpus based approach is mainly used in Statistical Machine Translation and the Example Based Machine Translation System.

### A. *Statistical Machine Translation*

Statistical Machine Translation is one of the most widely used machine translation approaches in the modern era. In Statistical Machine Translation, a bilingual corpus is trained and statistical parameters are derived in order to reach the most likely translation. Bilingual corpus are readily available for some languages, whereas others used methods like reputation based social collaboration [1] for building the corpora[20] or from web or any digitized text [21] etc.

Statistical Machine Translation takes place in three phases, namely language modeling, translation modeling and decoding.

The language model determines the probability of the target language T which helps in achieving the fluency in the target language and choosing the right word in the translated language. It is generally denoted as P (T).

The translation model, on the other hand helps to compute the conditional probability of the target language T given the source language S generally denoted as P (T|S).

Finally, in the decoding phase, the maximum probability of product of both the language model and the translation model is computed which gives the statistically most likely probable sentence in the target language.

$$P(S, T) = argmax\ P(T)\ P(S|T)$$

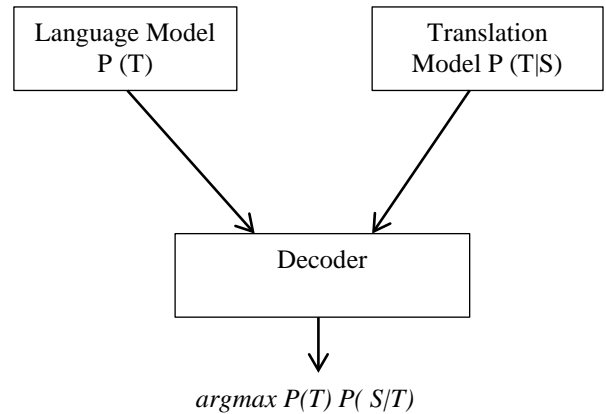

*argmax P(T) P( S|T)*

Figure 5 Statistical Machine Translation

The notable systems in Indian perspective which uses the Statistical Machine Translation approach are English Hindi Machine Translation system [14] developed by IIIT Hyderabad which combines Rule Based Machine Translation and phrase based Statistical Machine Translation approach. Another translation system was developed by Cochin University of Science and Technology for translating English to Malyalam [9] which used Statistical Machine translation approach.

### A. *Example Based Machine Translation System*

Example Based Machine Translation System usually uses previous translation examples to translate from source to target language. The basic idea of Example Based Machine Translation System is to retrieve examples of existing translation in its example-base and provide the new translation based on that example.

Example Based Machine Translation generally takes place in three phases – Matching, Alignment and Recombination. In the Matching phase, the system searches examples from the example-base that are similar to the input.

In the Alignment phase, the part of the example that is to be used is identified and aligned comparing with other examples. In the final phase, the reusable parts identified during the alignment phase are put together and produce the target language.

Machine Translation system in India employing the Example based approach includes Vaasaanubaada [4] is primarily targeted for Assamese Bengali bilingual translation of news text. Anubharti [8] developed at IIT Kanpur is a hybrid Example based system which combines pattern based and example based approach. Another notable system which uses the example based approach is the Shiva Machine Translation system [14] developed by IIIT Hyderabad and Carnegie Mellon University, USA.

## VI. MACHINE TRANSLATION SYSTEMS

The various Machine Translation systems for Indian languages with their language pair and features are given in Table 1:

*Table 1   Machine Translation Systems*

| TRANSLATION SYSTEM | APPROACH | LANGUAGE PAIR | FEATURES |
|---|---|---|---|
| Anusaaraka | Direct | Bengali, Kannada, Marathi, Punjabi, Telegu to Hindi | Uses Paninian grammar and matches local words between source and target language. |
| Hindi to Punjabi MTS | Direct | Hindi to Punjabi | Direct word to word translation. Morphological analysis, word sense disambiguation, post processing and transliteration |
| Mantra | Transfer Based | English to Hindi, Gujrati, Telegu. Hindi to English, Bengali, Marathi | Uses Tree Adjoining Grammar Formalism. |
| Matra | Transfer Based | English to Hindi | Human assisted translation project uses rule bases and heuristics. |
| Shakti | Transfer Based | English to Hindi, Marathi, Telegu | Works by combining rule based and statistical approach. |
| Anubaad | Transfer Based | English to Bengali | Hybrid system which uses n-gram approach for POS tagging. Works at sentence level |
| English Kannada MTS | Transfer Based | English to Kannada | Uses Universal Clause Structure Grammar Formalism |
| Sampark | Transfer Based | Punjabi-Hindi, Telegu to Tamil, Hindi-Urdu, Hindi to Telegu | Uses Computational Paninian Grammar (CPG) approach for analyzing language and combines it with machine learning |
| Anglabharti | Interlingual | English to Hindi, Tamil | Uses intermediate structure Pseudo Lingua for Indian Languages. |
| AnglaHindi | Interlingual | English to Hindi | Uses rule-bases, example-base and statistics to obtain translation for frequently encountered noun and verb phrasal. |
| UNL English Hindi MTS | Interlingual | English to Hindi | Uses Universal Natural Language as interlingua |
| English Hindi MTS | Statistical Machine Translation | English to Hindi | Combines Rule Based Machine Translation and phrase based Statistical Machine Translation |
| English Malyalam MTS | Statistical Machine Translation | English to Malyalam | Uses SMT by using monolingual Malayalam corpus and a bilingual English/Malayalam corpus in the training phase |
| Vaasaanubaada | Example Based Machine Translation | Bilingual Bengali Assamese | Preprocessing and post processing task, longer sentences fragmented at punctuation, backtracking for unmatched results. |
| Anubharti | Example Based Machine Translation | English-Hindi | Hybrid Example based system which combines pattern based and example based approach. |
| Shiva | Example Based Machine Translation | English- Hindi | Uses linguistic rules and statistical approach to infer linguistic information. |

## VII. CONCLUSION

In this paper, we discussed and looked at the various Machine Translation Systems in India along with their features. We also discussed the various approaches that are applied for building a machine translation system. Many researchers and research groups have come up with different translation systems applying different approaches. Though many languages are yet to be covered under Machine Translation in India, significant research and work has been going on to include many of the languages. Many languages and it is believed that in a few years' time most of the major Indian languages will be covered under the Machine translation ambit.